\pdfoutput=1

\documentclass[letterpaper, 10pt, conference]{ieeeconf}
\IEEEoverridecommandlockouts
\usepackage{cite}
\usepackage{amsmath,amssymb,amsfonts}
\usepackage{mathtools}
\usepackage{dsfont}
\usepackage{siunitx}
\usepackage{xfrac}
\usepackage{graphicx}

\usepackage{booktabs}
\usepackage{multirow}
\usepackage{makecell}

\usepackage{tikz}

\makeatletter
\let\NAT@parse\undefined
\makeatother
\usepackage[hidelinks]{hyperref}

\title{\LARGE \textbf{Simultaneous Clutter Detection and Semantic Segmentation of\\Moving Objects for Automotive Radar Data}}

\author{Johannes Kopp\textsuperscript{1}, Dominik Kellner\textsuperscript{2}, Aldi Piroli\textsuperscript{1}, Vinzenz Dallabetta\textsuperscript{2} and Klaus Dietmayer\textsuperscript{1}%
\thanks{\textsuperscript{1}Institute of Measurement, Control and Microtechnology, Ulm University, Albert-Einstein-Allee 41, 89081 Ulm, Germany {\tt\small \{firstname\}.\{lastname\}@uni-ulm.de}}%
\thanks{\textsuperscript{2}BMW AG, Petuelring 130, 80809 Munich, Germany \mbox{{\tt\small \{dominik.m.kellner, vinzenz.dallabetta\}@bmw.de}}}%
}

\newcommand\copyrighttext{
	\footnotesize \textcopyright 2023 IEEE.  Personal use of this material is permitted.  Permission from IEEE must be obtained for all other uses, in any current or future media, including reprinting/republishing this material for advertising or promotional purposes, creating new collective works, for resale or redistribution to servers or lists, or reuse of any copyrighted component of this work in other works.}
\newcommand\copyrightnotice[1]{
	\begin{tikzpicture}[remember picture,overlay]
	\node[anchor=north,yshift=-15pt] at (current page.north) {\parbox{\dimexpr\textwidth-1.0cm}{#1}};
	\end{tikzpicture}
	\vspace{-10pt}
}

\begin{document}

\maketitle
\copyrightnotice{\copyrighttext}
\thispagestyle{empty}
\pagestyle{empty}

%%%%%%%%%%%%%%%%%%%%%%%%%%%%%%%%%%%%%%%%%%%%%%%%%%%%%%%%%%%%%%%%%%%%%%%%%%%%%%%%
\begin{abstract}

The unique properties of radar sensors, such as their robustness to adverse weather conditions, make them an important part of the environment perception system of autonomous vehicles.
One of the first steps during the processing of radar point clouds is often the detection of clutter, i.e. erroneous points that do not correspond to real objects.
Another common objective is the semantic segmentation of moving road users.
These two problems are handled strictly separate from each other in literature. The employed neural networks are always focused entirely on only one of the tasks. In contrast to this, we examine ways to solve both tasks at the same time with a single jointly used model.
In addition to a new augmented multi-head architecture, we also devise a method to represent a network's predictions for the two tasks with only one output value.
This novel approach allows us to solve the tasks simultaneously with the same inference time as a conventional task-specific model.
In an extensive evaluation, we show that our setup is highly effective and outperforms every existing network for semantic segmentation on the RadarScenes dataset~\cite{RadarScenesDataset}.

\end{abstract}

%%%%%%%%%%%%%%%%%%%%%%%%%%%%%%%%%%%%%%%%%%%%%%%%%%%%%%%%%%%%%%%%%%%%%%%%%%%%%%%%
\section{Introduction} \label{section:introduction}

Radar sensors play an important role in the environment perception for advanced driver assistance systems and autonomous vehicles. They have a large range of up to several hundred meters, are robust to adverse weather conditions such as rain or fog, and can directly measure the velocity of objects in a single cycle. Radars thus complement the information provided by cameras and lidar sensors and add further redundancy for safety-critical applications.

\begin{figure}[!t]
	\centering
	\vspace*{0.19cm}
	\includegraphics{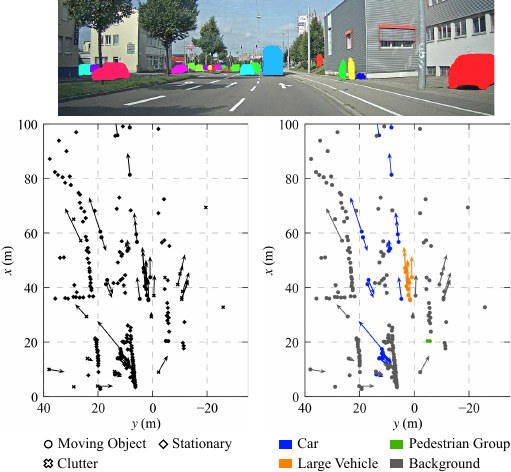}
	\caption{Example of a radar point cloud, where detections are classified with regard to either clutter detection (left) or semantic segmentation (right). Symbols mark the positions of detections relative to the ego vehicle, arrows visualize their velocity over ground. Colored patches in the camera image cover objects for privacy reasons and have no meaning~\cite{RadarScenesArxiv}.}
	\label{fig:example_tasks}
\end{figure}

In this work, we consider two common tasks within the context of the environment perception based on radar. The first is the detection of clutter in radar point clouds. At the end of every ``scan'', automotive radar sensors typically output a list of so-called detections or targets. Each of these detection points is meant to indicate the position and radial velocity of an object.
However, effects like incorrect ambiguity resolution during the sensor's signal processing, interference between sensors, and multipath propagation result in the occurrence of errors. Thus, many of the detections do not actually match any real object in the environment. Such clutter points are highly problematic for perception methods like object detection or tracking. It is therefore important to identify them, so that they can be removed before such steps. This is what is done by \textit{clutter detection}.
The left side of Fig.~\ref{fig:example_tasks} illustrates the task.

The second task we study is the \textit{semantic segmentation} of moving objects for radar point clouds. Here, each detection must be classified regarding what type of object it represents (e.g. car, pedestrian, etc.). Since the static environment is usually analyzed with specialized methods like occupancy grid mapping~\cite{Engels2021, Schumann2020, Prophet2020}, only moving objects have to be considered. All other detections, including clutter, count as background.
An example of a segmented point cloud is shown on the right side of Fig.~\ref{fig:example_tasks}.

In existing literature, clutter detection and semantic segmentation are strictly separated. Approaches always focus on only one of the two tasks without considering the other. The main differences are the high importance of analyzing relationships between points that lie far away from each other for clutter detection (cf.~\cite{Kopp2021}), and that clutter detection, which often acts as a preprocessing step, must have a very low execution time.
However, we notice that the tasks are still very similar in terms of the available input data and the neural network architectures that are commonly used for tackling them. We therefore explore ways to solve both tasks simultaneously with the same network.
Our main contributions are the following:
\begin{itemize}
	\item We design a multi-head network that is capable of detecting clutter and performing a semantic segmentation of radar point clouds at the same time. A novel post-processing module guarantees consistency between the predictions and improves the performance.
	\item We show how the class definitions of the two tasks can be fused into a single label.
	This allows us to solve the tasks simultaneously also with a normal single-head network.
	\item We compare the new architectures with a reference setup in which clutter detection and semantic segmentation are performed one after the other, and with approaches that are focused entirely on just one of the two tasks.
	Our best setup solves both tasks simultaneously without any increase of the inference time compared to a specialized network. On top of that, it outperforms every existing architecture for semantic segmentation on the popular RadarScenes dataset~\cite{RadarScenesDataset}.
\end{itemize}

%%%%%%%%%%%%%%%%%%%%%%%%%%%%%%%%%%%%%%%%%%%%%%%%%%%%%%%%%%%%%%%%%%%%%%%%%%%%%%%%
\section{Related Work} \label{section:related_work}

The fundamentals of clutter in automotive radar data are extensively studied e.g. in \cite{Kopp2021, Kamann2018, Holder2019}. The latter gives a broad overview of possible causes.
The most common cases of multipath propagation in road traffic and how they are reflected in the final point cloud are analyzed in \cite{Kopp2021}. For the specular reflection at walls or guardrails, even more details and results of practical experiments can be found in \cite{Kamann2018}.

Regarding the actual detection of clutter using neural networks, several approaches exist. Many of them are limited to identifying only certain types of errors, however.
For example, clutter resulting from specular reflections is searched using PointNet++~\cite{PointNet++} or a Similarity Group Proposal Network~\cite{SGPN} (SGPN) in \cite{Kraus2020} and \cite{Kraus2021}, respectively.
The datasets utilized in both works are restricted to a controlled environment and a stationary ego vehicle. Other approaches designed for detecting certain kinds of clutter are \cite{Wang2021} and \cite{Griebel2021}. The latter extends PointNet++ by a new grouping mechanism created specifically for this task.
A more general work is e.g. \cite{Jin2021}. The authors compare a random forest, a convolutional neural network (CNN) and PointNet++ for the detection of arbitrary clutter. But they use a very small dataset and consider only points that are extremely close to the sensor. The only works without such limitations are \cite{Chamseddine2020} and \cite{Kopp2023}.
The former is based on PointNet~\cite{PointNet} and uses proprietary data, hindering the comparison with other approaches. In the latter, we generate ground truth for a public radar clutter dataset and present a new PointNet++ setup for clutter detection.

Literature regarding the semantic segmentation of radar point clouds is more homogeneous.
Approaches are targeted on the classification of either static or dynamic objects. The analysis of the static environment is typically based on the creation of occupancy grid maps, which are continually updated with the newest stationary points.
The grid cells are then classified using a CNN.
Examples of this can be found in \cite{Schumann2020, Prophet2020, Lombacher2017}.
Regarding the semantic segmentation of moving objects, the release of the large high-quality RadarScenes dataset~\cite{RadarScenesDataset} has led to an influx of research.
Here, most works employ PointNet++ as their base architecture. For example, \cite{Schumann2018} presents a setup to directly leverage the original network for radar processing. This setup is further extended with an internal memory for recursively storing data of previous time steps in \cite{Schumann2020}. The stored information is used to calculate additional point features and improve segmentation accuracy. Other modifications of PointNet++, such as replacing the sampling step with mean shift clustering or adding more processing blocks, are described in \cite{Cennamo2020} and \cite{Liu2023}.
Approaches for the semantic segmentation of moving objects that are not based on PointNet++ are \cite{Zeller2023} and \cite{Fent2023}.
Instead, an adapted PointTransformer architecture~\cite{PointTransformerZhao} and a graph neural network are used, respectively.

Clutter detection and semantic segmentation are conducted strictly separate from each other in literature. While some of the setups for clutter detection have been inspired by segmentation approaches, no authors have previously attempted to jointly solve both tasks with a single network.
The effect of a preceding removal of clutter on segmentation models has also not been investigated. We address both of these issues in this work.

%%%%%%%%%%%%%%%%%%%%%%%%%%%%%%%%%%%%%%%%%%%%%%%%%%%%%%%%%%%%%%%%%%%%%%%%%%%%%%%%
\section{Dataset} \label{section:dataset}

We use the public RadarScenes dataset~\cite{RadarScenesDataset} for all experiments and exemplary point cloud visualizations presented in this paper. It contains a total of \SI{119}{M} radar detections recorded during \SI{4.3}{h} of driving.
Sequences include urban traffic, rural routes and scenarios where the ego vehicle is standing still. Four series-production automotive 2D \SI{77}{GHz} radar sensors are employed.
Each sensor has a range of about \SI{100}{m} and covers a horizontal angle of \ang{120}. The sensors face toward the front and sides of the vehicle with overlapping fields of view.

The ground truth of RadarScenes is aimed at the semantic segmentation of moving objects. With the recommended official configuration, the six distinguished classes are \textit{car}, \textit{pedestrian}, \textit{pedestrian group}, \textit{two-wheeler}, \textit{large vehicle} and \textit{background}\footnote{The background class is called ``static'' in the dataset even though it also includes detections with high velocities (e.g. clutter). To prevent confusion, we refer to it as ``background'' throughout this paper.}.
Each detection point whose position perfectly matches that of a moving object is labeled with the respective class. The one exception to this are detections which belong to highly unusual object types like animals or skaters.
These are not annotated and should be excluded from loss calculation and evaluation~\cite{RadarScenesArxiv}.
All remaining detections, including those with slight measurement errors and clutter, are marked as background.

\begin{figure}[!t]
	\centering
	\vspace*{0.15cm}
	\includegraphics{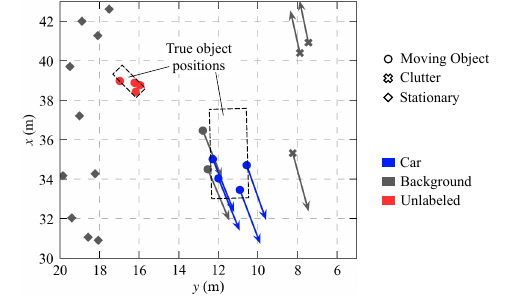}
	\caption{Illustrative scene in which the labels for clutter detection and semantic segmentation seemingly contradict each other.
	Both tasks' annotations are shown in the same plot.
	Detections that stem from an object but do not accurately reflect its size due to slight positional errors are assigned to the classes \textit{moving object} and \textit{background}, respectively (cf. center). They do not count as clutter.
	Detections belonging to very rare object types are marked as \textit{moving object} but remain unlabeled \mbox{regarding segmentation (cf. left).}}
	\label{fig:example_contradictory_labels}
\end{figure}

To obtain additional annotations regarding clutter detection, we apply the label generation method presented in \cite{Kopp2023}. With it, a second label for each point can be determined. Based on the original segmentation ground truth and the positions and velocities of detections, the data is divided into three new classes. Detections that correspond to any kind of object in motion are annotated as \textit{moving object}.
The other detections are split into those stemming from nonmoving objects, labeled as \textit{stationary}, and \textit{clutter}.
Unlike for semantic segmentation, small ordinary measurement errors are tolerated. This means that detections do not have to lie perfectly inside an object's bounding box to be included in the corresponding class.
Furthermore, all types of objects are considered. There are thus two cases in which points count as \textit{moving object} regarding clutter detection but are not part of any object class for semantic segmentation. These are visualized in Fig.~\ref{fig:example_contradictory_labels}.

%%%%%%%%%%%%%%%%%%%%%%%%%%%%%%%%%%%%%%%%%%%%%%%%%%%%%%%%%%%%%%%%%%%%%%%%%%%%%%%%
\section{Methods} \label{section:methods}

\begin{figure*}[!t]
	\centering
	\vspace*{0.18cm}
	\includegraphics{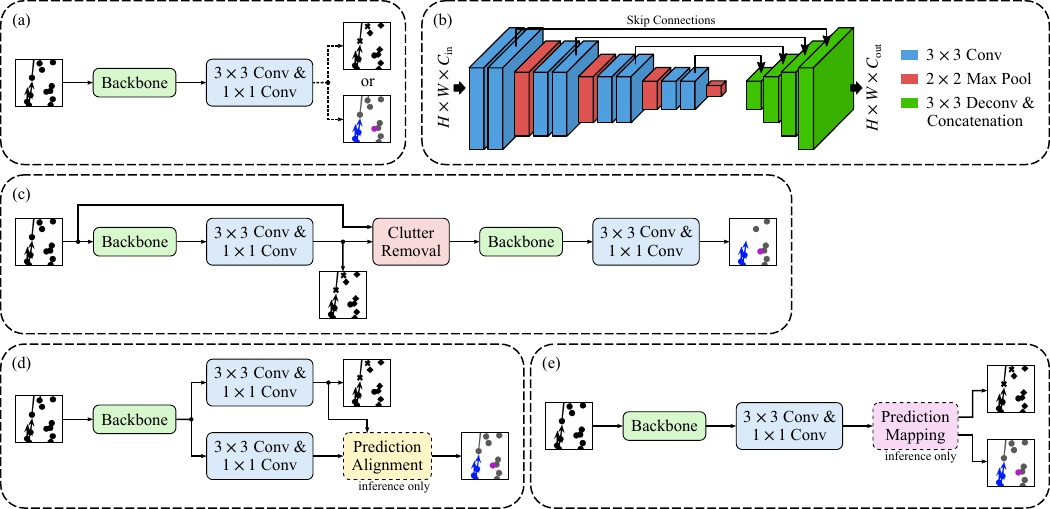}
	\caption{Overview of examined network architectures.
	(a) Single-task network that can be used either for clutter detection or semantic segmentation.\linebreak
	(b) Employed backbone architecture. Max pooling and deconvolutions use a stride of $2$, and thus halve and double the dimensions of grid maps, respectively.
	(c) Series connection of single-task models. Detections predicted to be clutter are removed from the data before the segmentation.
	(d) Multi-head network. The prediction alignment module ensures consistency between the predictions during inference and improves the performance.
	(e) Label fusion approach. The network first classifies points regarding fused labels. This output is then converted to the corresponding task-specific predictions during inference.
	}
	\label{fig:network_architectures}
\end{figure*}

\subsection{Basic Network Setup} \label{subsection:base_setup}

The detection of clutter and the semantic segmentation of moving objects can be arranged in the processing chain in several ways. The simplest option is to solve the tasks independently from each other with two separate neural networks. This approach is the state of the art in literature. Therefore, we design a basic network setup that can be used for performing either of the two tasks. It acts both as the basis for our new approaches in the following sections and as a reference during evaluation. In principle, any architecture capable of classifying individual points in a point cloud can be employed. We decide to use a CNN.
Our custom setup is simple, very fast and still manages to achieve highly competitive performance. An overview of the network is given in Fig.~\ref{fig:network_architectures}(a).

The first step of the CNN setup is the preparation of input data. Scans of all four radar sensors mounted on the test vehicle are accumulated over a sliding time window of \SI{150}{ms}. To combine them into a single point cloud, the positions of all detections are transformed to the current Cartesian vehicle coordinate system. This process increases the density of points and gives the network access to temporal information.
Since CNNs require an image-like representation of data, the resulting 2D point clouds are then converted to bird's-eye-view grid maps. Each detection's features are transferred to the grid cell it lies in. If a cell contains more than one detection, it uses the maximum of each feature. Empty cells keep their initial values of zero.
It should be noted that this procedure is not the same as generating a (dynamic) occupancy grid map, as the map is not recursively updated. Instead, an entirely new grid map is created whenever a sensor outputs a scan.

Once the preparation of input data is completed, the resulting grid map is processed by the network. We employ the optimized shallow U-Net~\cite{UNet} architecture shown in Fig.~\ref{fig:network_architectures}(b) as backbone. Any other suitable structure, such as a ResNet~\cite{ResNet}, could also be used.
Lastly, each cell's prediction is transferred back to all detections it contains. This yields the final pointwise classifications.

The described setup can be trained either for clutter detection or for semantic segmentation, depending on the used labels. The models obtained this way are always specific to just one of the two tasks, however. If both tasks should be solved by a practical system, two separate models must be trained and executed. This results in two times the number of computations compared to a single network. When there are not enough resources for parallel execution, the delay introduced into the processing chain is also doubled.

\subsection{Series Connection of Single-Task Models}

As mentioned in Sec.~\ref{section:introduction}, clutter detection is usually employed as a preprocessing step before other perception methods. Accordingly, a natural approach when using two separate models for clutter detection and semantic segmentation is to perform the tasks one after the other. This sequential processing allows the segmentation network to benefit from the results of clutter detection. An intuitive method for this is to remove all detections that were predicted to be clutter from the input point cloud.
We also found this to be more effective than e.g. using the class scores predicted by the first model as additional input features for the segmentation.
A schematic of the approach is given in Fig.~\ref{fig:network_architectures}(c).

We implement the series connection of clutter detection and semantic segmentation as a reference for our novel multi-task setups. Both of the models use the architecture described in the previous section. Since they expect grid maps of the same form as input, the removal of clutter can be accomplished efficiently by setting the corresponding grid cells to zero. Beyond the preparation of input data, there are no shared processing steps.
The total computational complexity and inference time of the stack are thus two times that of a single network.

\subsection{Multi-Head Network}

The first setup we design that is capable of performing clutter detection and semantic segmentation simultaneously is based on a multi-head network. Following a shared backbone, the architecture is split into two parallel branches with heads for the individual tasks. Fig.~\ref{fig:network_architectures}(d)  visualizes the structure.

Usually, the outputs of the different heads of a multi-head network are not coordinated.
In our case, this can lead to the situation that the two predictions are incompatible with each other. A point might be classified as \textit{clutter} or \textit{stationary} by the head for clutter detection, while the segmentation head predicts an object class. To prevent this from happening, we introduce a novel post-processing step that aligns the predictions. For all detections that are classified as \textit{clutter} or \textit{stationary}, the output of the segmentation head is overwritten and automatically set to \textit{background}. Since this operation is not differentiable and would prevent the backpropagation of gradients for the affected points, it is applied only during inference. In addition to guaranteeing consistency between predictions, the new post-processing step results in an even heavier exploitation of the synergies between the two tasks and improves the segmentation accuracy.

It is worth noting that, even with the prediction alignment activated, a detection can still be classified as both \textit{moving object} and \textit{background}. This must be allowed due to the different treatment of small measurement errors in the tasks' ground truths (cf. Sec.~\ref{section:dataset}).
Compared to the basic network setup from Sec.~\ref{subsection:base_setup}, the multi-head architecture performs clutter detection and semantic segmentation simultaneously at the sole cost of adding two convolutional layers and the post-processing module.
The computational complexity and the inference time are only moderately increased.

\subsection{Label Fusion}  \label{subsection:label_fusion}

In addition to the multi-head network, we devise a second, even more efficient method for performing clutter detection and semantic segmentation at the same time. It is based on the idea of combining the ground truths of the two tasks into a single label. In general, covering every possible pairing of the labels from three and seven categories (counting also unlabeled detections) requires $3 \cdot 7 = 21$ new classes.
Distinguishing that many types would only be possible with a large and slow model. However, in our particular case, the specific properties of the two tasks can be leveraged to reduce the necessary number of classes.
Points which are assigned to an object type or remain unlabeled regarding semantic segmentation never belong to the classes \textit{clutter} or \textit{stationary} of clutter detection (cf. Sec.~\ref{section:dataset}).
As a result, $6 \cdot 2 = 12$ of the new classes can be eliminated. This leaves only $9$ to be distinguished. In particular, we adapt the automatic label generation method of clutter detection~\cite{Kopp2023} to instead produce more finely differentiated fused labels.
To the original five object types of RadarScenes, we add the classes \textit{other object}, \textit{inaccurate measurement}, \textit{clutter} and \textit{stationary}. \textit{Other object} identifies the detections that are not annotated for semantic segmentation. It thus contains all points perfectly matching the position of any moving object that is not covered by the first five classes. Detections which correspond to an object but, due to small measurement errors, do not correctly represent the object's dimensions, are marked as \textit{inaccurate measurement}. For finding them, the same set of rules as during the generation of clutter labels is employed (as described in \cite{Kopp2023}). The other classes keep their previous definitions.

\begin{figure}[!t]
	\centering
	\vspace*{0.18cm}
	\includegraphics{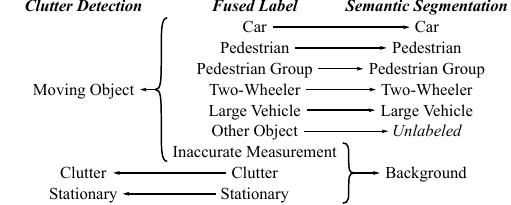}
	\caption{Label mappings to extract ground truths of clutter detection and semantic segmentation from the fused labels}
	\label{fig:label_mapping}
\end{figure}

The fused labels that are obtained via our new generation method contain all information of the ground truths of both clutter detection and semantic segmentation. By applying a simple mapping, the annotations of either task can perfectly be restored. The mapping function is visualized in Fig.~\ref{fig:label_mapping}.

Representing the two tasks with only a single label has a huge advantage.
It makes it possible to solve both tasks simultaneously using a normal single-head network.
When one or both of the task-specific classifications are desired during inference, the predictions of the network can simply be mapped in the same way as the labels. This setup is shown in Fig.~\ref{fig:network_architectures}(e). The approach is highly efficient. As the architecture is virtually identical to that of a model performing just one of the tasks, the computational complexity and inference time also remain nearly unchanged.

For evaluating the label fusion approach, we use the same CNN setup as in the other sections. However, we decide to configure the network to distinguish only seven of the nine new classes. The classes \textit{other object} and \textit{inaccurate measurement}, which are necessary to preserve all information of the individual tasks' labels but occur only very infrequently, are not considered by the model.
The corresponding detections can thus be excluded from loss calculation, which balances the class distribution and further eases training. On the flip side, this also means that the network is forced to misclassify the affected detections regarding at least one of the individual tasks. Nonetheless, our experiments show that the approach is overall beneficial and improves results (see Sec.~\ref{subsection:exclusion_rare_classes}).

%%%%%%%%%%%%%%%%%%%%%%%%%%%%%%%%%%%%%%%%%%%%%%%%%%%%%%%%%%%%%%%%%%%%%%%%%%%%%%%%
\section{Experiments and Results} \label{section:experiments}

{
\begin{table*}[!t]
	\centering
	\vspace*{0.18cm}
	\caption{Comparison of network setups for clutter detection and/or semantic segmentation. Some of the approaches are recreated from their description in literature to test them on the RadarScenes dataset or to determine their inference time.
	Macro-averaged performance values are given in \si{\percent}.
	We measure all inference times, i.e. the average times networks require for processing one point cloud, on an Nvidia RTX 2080 Ti GPU. Schumann et al. use an Nvidia GTX 1080 in \cite{Schumann2020} instead.
	}
	\label{tab:results_comparison}
	\begin{tabular}{ c c | c c c | c c c | c }
		\toprule
		\multirowcell{2}{Approach} & \multirowcell{2}{Remarks} & \multicolumn{3}{c|}{Clutter Detection} & \multicolumn{3}{c|}{Semantic Segmentation} & \multirowcell{2}{Inference Time}\\
		&& Precision & Recall & F1 Score & Precision & Recall & F1 Score &\\
		\hline\rule{0pt}{9pt}%
		Ours, single-task network && $92.46$ & $\mathbf{95.50}$ & $93.92$ & -- & -- & -- & \textbf{\SI[detect-weight=true]{6.2}{ms}}\\
		Ours, single-task network && -- & -- & -- & $75.56$ & $\mathbf{84.34}$ & $79.41$ & \textbf{\SI[detect-weight=true]{6.2}{ms}}\\
		Ours, series connection of models && $92.46$ & $\mathbf{95.50}$ & $93.92$ & $79.42$ & $81.97$ & $80.56$ & \SI{11.5}{ms}\\
		Ours, multi-head network && $91.96$ & $\mathbf{95.50}$ & $93.63$ & $78.51$ & $84.11$ & $81.04$ & \SI{8.0}{ms}\\
		Ours, label fusion approach && $\mathbf{93.06}$ & $95.43$ & $\mathbf{94.21}$ & $\mathbf{79.90}$ & $83.92$ & $\mathbf{81.78}$ & \textbf{\SI[detect-weight=true]{6.2}{ms}}\\
		\hline\rule{0pt}{9pt}%
		Kraus et al. \cite{Kraus2020} & Reimplementation & $74.68$ & $89.95$ & $80.93$ & -- & -- & -- & \SI{6.6}{ms}\\
		Griebel et al. \cite{Griebel2021} & Reimplementation & $91.47$ & $92.64$ & $92.00$ & -- & -- & -- & \textbf{\SI[detect-weight=true]{4.0}{ms}}\\
		Kopp et al. \cite{Kopp2023} && $\mathbf{94.00}$ & $\mathbf{96.11}$ & $\mathbf{95.03}$ & -- & -- & -- & \SI{7.0}{ms}\\
		Kopp et al. \cite{Kopp2023} w/o accumulation && $92.77$ & $94.41$ & $93.55$ & -- & -- & -- & \SI{5.7}{ms}\\
		\hline\rule{0pt}{9pt}%
		Schumann et al. \cite{Schumann2018} & As reported in \cite{SchumannThesis} & -- & -- & -- & $73.9$ & $82.1$ & $77.6$ & n/a\\
		Schumann et al. \cite{Schumann2018} & Reimplementation & -- & -- & -- & $72.82$ & $80.24$ & $75.92$ & \textbf{\SI[detect-weight=true]{7.7}{ms}}\\
		Schumann et al. \cite{Schumann2020} & As reported in \cite{SchumannThesis} & -- & -- & -- & $\mathbf{77.7}$ & $\mathbf{84.9}$ & $\mathbf{81.1}$ & \SI{100}{ms}\\
		Zeller et al. \cite{Zeller2023} && -- & -- & -- & n/a & n/a & $79.8$ & n/a\\
		Fent et al. \cite{Fent2023} && -- & -- & -- & n/a & n/a & $77.1$ & n/a\\
		\bottomrule
	\end{tabular}%
\end{table*}
}

{
\setlength{\tabcolsep}{1.84mm}
\begin{table*}[!t]
	\centering
	\caption{Class-specific F1 scores (in \si{\percent}) regarding clutter detection and/or semantic segmentation achieved by our network setups}
	\label{tab:class_specific_performance}
	\begin{tabular}{ c | c c c | c c c c c c }
		\toprule
		\multirowcell{2}{Approach} & \multicolumn{3}{c|}{Clutter Detection} & \multicolumn{6}{c}{Semantic Segmentation}\\
		& Moving Obj. & Clutter & Stationary & Car & Pedestrian & Ped. Group & 2-Wheeler & Large Vehicle & Background\\
		\hline\rule{0pt}{9pt}%
		Single-task network & $89.81$ & $92.32$ & $99.64$ & -- & -- & -- & -- & -- & --\\
		Single-task network & -- & -- & -- & $83.39$ & $52.87$ & $80.86$ & $80.93$ & $78.85$ & $99.55$\\
		Series connection of models & $89.81$ & $92.32$ & $99.64$ & $85.31$ & $53.67$ & $82.06$ & $81.50$ & $81.17$ & $\mathbf{99.65}$\\
		Multi-head network & $89.36$ & $91.91$ & $99.62$ & $85.79$ & $54.88$ & $81.88$ & $82.73$ & $81.30$ & $99.64$\\
		Label fusion approach & $\mathbf{90.45}$ & $\mathbf{92.51}$ & $\mathbf{99.68}$ & $\mathbf{86.34}$ & $\mathbf{55.93}$ & $\mathbf{82.75}$ & $\mathbf{84.23}$ & $\mathbf{81.77}$ & $\mathbf{99.65}$\\
		\bottomrule
	\end{tabular}%
\end{table*}
}

\subsection{Setup Details}

For evaluating the network architectures proposed in the previous sections, we set the cell size of grid maps fed to the CNN backbones to \SI{40}{cm}. This value strikes a balance between accuracy and computational cost. Each grid map consists of $512 \times 512$ cells with the ego vehicle located slightly below the center, and covers the entire area of $x \in [\SI{-95.2}{m}, \SI{109.6}{m})$, $y \in [\SI{-102.4}{m}, \SI{102.4}{m})$ in which detections occur. The individual cells are described by the position, the radar cross-section and the ego motion compensated velocity of the detections they contain.
The velocity is specified as a vector with $x$- and $y$-component.
We deliberately keep the backbone architecture slim. Convolutional layers are configured to alternate between $64$ and $32$ output channels. Deconvolutions always output $32$ channels.

All of our network setups are trained using the same settings.
The training configuration is largely identical to one for clutter detection we describe in \cite{Kopp2023}. Most importantly, we train for $20$ epochs using Adam optimization, a cyclical learning rate policy and focal loss.
The class weights for loss calculation regarding clutter detection and semantic segmentation are determined as in \cite{Kopp2023} and \cite{SchumannThesis}, respectively.
For the multi-head network, the two task-specific loss terms are averaged to obtain the total loss.
The label fusion approach is trained not regarding the individual tasks but directly with respect to the fused labels. Here, we empirically set the class weights to $0.70$ for stationary detections, $3.52$ for clutter and $4.93$ for the remaining classes.
Finally, the series connection of networks is trained in a two-stage process. First, the front model is optimized for clutter detection. Then its layers are frozen, the segmentation model is appended and only this model is trained.

For evaluation, we repeat the training of each network setup five times to minimize random influences.
The average performance of those runs on the validation set is then reported. Metrics incorporate only predictions for the most recent sensor scan in the point cloud. Older detections have already been classified in a previous time step and are added solely for context (see Sec.~\ref{subsection:base_setup}). When evaluating the series connection of single-task models regarding segmentation, points that are removed for being clutter are counted as if the network classified them as \textit{background}.
This enables a fair comparison with other approaches.

\subsection{Comparison of Different Approaches} \label{subsection:results_comparison}

We evaluate the presented network setups regarding their inference time and their performance for clutter detection and the semantic segmentation of moving objects. The networks are compared not just with each other but also with approaches in literature that are focused on only one of the two tasks. The results of the experiments are listed in Tab.~\ref{tab:results_comparison}.

The comparison shows that all of our setups reach a performance that is similar to or even exceeds the respective state of the art in literature.
This is true even for the approaches that solve both considered tasks at the same time. The inference times are well below a quarter of the radar sensors' cycle length of \SI{60}{ms}. All networks are hence suited for real-time execution and can directly be deployed for practical applications.
The series connection of models of course has the same performance regarding clutter detection as the corresponding single-task network. But it is more likely to miss detections belonging to object classes than the specialized segmentation network, which is reflected in a lower class-averaged recall.
This is caused by the erroneous removal of detections that were misclassified as clutter by the first model.
The clutter removal also leads to a considerable increase of the precision and thus the F1 score, though.
Our new multi-head network is nearly as good at detecting clutter as the single-task setup. At the same time, it also performs a semantic segmentation. There, it achieves even better results than the model trained exclusively for the task.
Most of the improvement comes from the novel prediction alignment module, which helps to prevent background detections from being misclassified as objects. This results in a large increase of precision and F1 score. Furthermore, the multi-head approach is much faster than using two separate networks. Compared to the basic single-task setup, the inference time increases by only about \SI{29}{\percent}.
The label fusion approach even surpasses these results. It outperforms the single-task models also regarding clutter detection. Concerning segmentation, the simultaneous use of backbone and head also for identifying clutter leads to an even better precision than before.
The label fusion approach therefore reaches the highest F1 score of all our setups for both clutter detection and semantic segmentation. All of this is achieved without any increase of the inference time. The network takes just as long for solving both tasks as the specialized models take for only a single one.

\begin{figure*}[!t]
	\centering
	\vspace*{0.18cm}
	\includegraphics{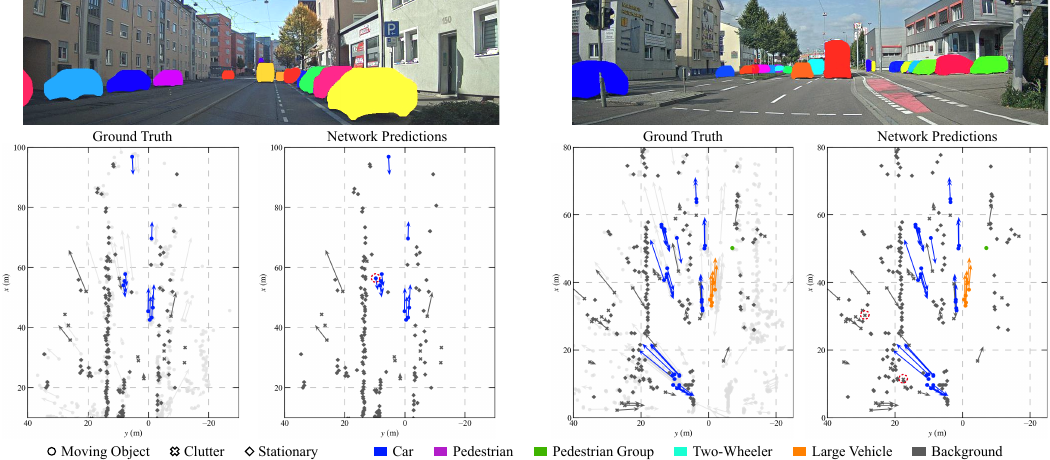}
	\caption{Exemplary data samples and the corresponding predictions of a network implementing our label fusion approach.
	Symbols represent the classes of clutter detection, colors indicate segmentation classes.
	Light gray points\includegraphics{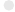} in the ground truth visualizations belong to old scans in the input point cloud fed to the model. The respective predictions are not relevant and thus not drawn.
	Incorrect predictions are highlighted by red circles.
	}
	\label{fig:example_predictions}
\end{figure*}

Compared with network designs in literature, which always focus entirely on just one task, our label fusion approach represents a substantial advancement.
Regarding clutter detection, its performance exceeds that of most setups and is only slightly below the state of the art. The difference is compensated for with a faster execution and the second output.
Regarding semantic segmentation, our approach manages to achieve the highest F1 score that has ever been reported. It outperforms every existing network in literature and sets a new record on the RadarScenes dataset. On top of that, its inference time is less than $\sfrac{1}{10}$ of that of the former best setup \cite{Schumann2020}.
This means that generating fused labels and using them for training is advantageous even if the network output for clutter detection is not needed at all. The procedure increases performance without affecting the inference time.
By differentiating the classes of labels more finely than is usual for the individual tasks, the detections within each class become more uniform in their characteristics.
This seems to make it easier for a model to learn to distinguish classes, which benefits both tasks.
Some visual examples of the predictions of our network are shown in Fig.~\ref{fig:example_predictions}.

\subsection{Class-Specific Performance of Setups}

To enable an even more detailed assessment of the quality of predictions produced by our setups, their performances regarding individual classes are given in Tab.~\ref{tab:class_specific_performance}.
As can be seen, the coarse magnitude of F1 scores is the same for all networks. The highest value for each class is achieved by the label fusion approach. Independent of the considered task, the category to which the majority of detections belong, i.e. \textit{stationary} or \textit{background}, is also the easiest to identify for the networks.
The class with the lowest performance values by far is \textit{pedestrian}.
This is because the recognition is often impeded by the sensors outputting only a single detection, which has a very low velocity, for each person.

\begin{figure}[!b]
	\centering
	\includegraphics{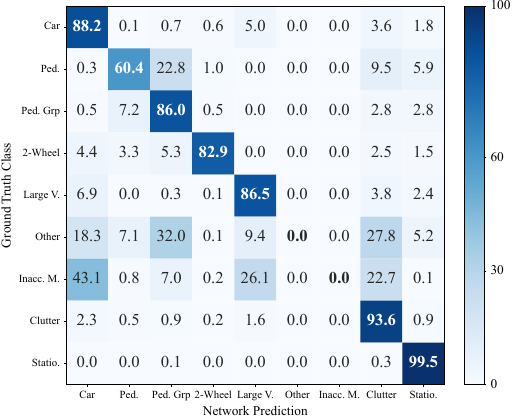}
	\caption{Confusion matrix of network trained with fused labels before mapping the predictions to any individual task. All values are given in \si{\percent}.}
	\label{fig:confusion_matrix_ext_sem_seg}
\end{figure}

To better understand the nature of errors made by the networks, we analyze the confusion matrix of the label fusion approach. The relationships between ground truth classes and the respective predictions before the task-specific mappings are presented in Fig.~\ref{fig:confusion_matrix_ext_sem_seg}. The matrix reveals that the comparatively low performance for pedestrians mainly stems from a frequent misclassification as \textit{pedestrian group}.
The two classes are quite similar in their characteristics, especially regarding the velocity and radar cross-section of associated detections.
As objects of both types require the same behavior of the ego vehicle, their confusion is tolerable.
A similar relationship exists between cars and large vehicles, which are primarily mistaken for each other. Due to the smooth transition between very large cars, such as SUVs, and small vans or trucks, the classes cannot be cleanly separated.
As described in Sec.~\ref{subsection:label_fusion}, we configure our network to distinguish only seven of the nine classes of fused labels.
For detections annotated as \textit{other object}, the first class that cannot be predicted, the model outputs are instead spread somewhat evenly over the remaining categories.
The most common predictions are \textit{pedestrian group} and \textit{clutter}. This is presumably because objects assigned to the class are typically small and have unusual characteristics (e.g. skaters or a person with a pushcart).
Object detections with slightly inaccurate positions are mostly predicted to be cars or large vehicles. These are the object types for which the corresponding measurement errors occur most often.

\subsection{Exclusion of Rare Classes in the Label Fusion Approach} \label{subsection:exclusion_rare_classes}

In our main setup for the label fusion approach, models are configured not to distinguish the classes \textit{other object} and \textit{inaccurate measurement}.
We decide to ease the learning process through the exclusion of those labels at the cost of forcing the network to misclassify the corresponding detections.
To quantify the effect on the final performance, we also try training with all nine fused labels (see Tab.~\ref{tab:results_label_fusion_all_classes}). In the first tested version (v1), the loss weights for the newly added classes are the same as for object types.
The resulting performance is only slightly below that of the seven-class network. This is because the model learns to almost never predict the new categories. To circumvent this behavior, classes are weighted proportionately to their frequencies in the second version of the setup (v2). But the network still does not manage to reach acceptable accuracies for the additional labels. As a result, the overall performance decreases significantly compared to the setup with seven distinguished classes, particularly regarding segmentation.

{
\begin{table}[!t]
	\centering
	\vspace*{0.18cm}
	\caption{Macro-averaged performance (in \si{\percent}) of label fusion approach with networks distinguishing different numbers of classes
	}
	\label{tab:results_label_fusion_all_classes}
	\begin{tabular}{ c | c c c | c c c }
		\toprule
		\multirowcell{2}{\# Distin.\\Classes} & \multicolumn{3}{c|}{Clutter Detection} & \multicolumn{3}{c}{Semantic Segmentation}\\
		& Prec. & Recall & F1 & Prec. & Recall & F1\\
		\hline\rule{0pt}{9pt}%
		$7$ & $\mathbf{93.06}$ & $\mathbf{95.43}$ & $\mathbf{94.21}$ & $\mathbf{79.90}$ & $83.92$ & $\mathbf{81.78}$\\
		$9$ (v1) & $92.78$ & $95.37$ & $94.03$ & $79.25$ & $83.98$ & $81.45$\\
		$9$ (v2) & $91.89$ & $95.01$ & $93.34$ & $74.99$ & $\mathbf{84.00}$ & $79.00$\\
		\bottomrule
	\end{tabular}%
\end{table}
}

%%%%%%%%%%%%%%%%%%%%%%%%%%%%%%%%%%%%%%%%%%%%%%%%%%%%%%%%%%%%%%%%%%%%%%%%%%%%%%%%
\section{Conclusion} \label{section:conclusion}

In this work, we present four different strategies to perform clutter detection and a semantic segmentation of moving objects for radar point clouds. The state-of-the-art approach is to use two separate neural networks, each tailored specifically to one of the tasks. We investigate the effect of executing these networks one after the other and removing clutter from the input to the segmentation.
Furthermore, we design two novel network setups that are capable of solving both considered tasks simultaneously with only a single model.
Our multi-head network uses a new prediction alignment module to coordinate the outputs of its two heads.
In our label fusion approach, the class definitions of clutter detection and semantic segmentation are combined. This allows us to jointly solve the tasks also with a normal single-head architecture. Both of our setups achieve performance values similar to those of networks focused entirely on only one of the two tasks. The label fusion approach even surpasses the corresponding single-task models.
With a mean F1 score of \SI{81.78}{\percent} on the RadarScenes dataset, it outperforms every existing setup in literature regarding the semantic segmentation.
All of this is accomplished without any increase of the inference time compared to a specialized network.

%%%%%%%%%%%%%%%%%%%%%%%%%%%%%%%%%%%%%%%%%%%%%%%%%%%%%%%%%%%%%%%%%%%%%%%%%%%%%%%%%
\bibliographystyle{IEEEtran}
\bibliography{references}

\end{document}